\documentclass[11pt]{article}

\usepackage{amsmath}
\usepackage{amsthm}
\usepackage{graphicx}
\usepackage{algpseudocode}
\usepackage{algorithm}
\usepackage{amsthm}

\usepackage{lastpage}

\begin{document}

\title{Minimum number of neurons in fully connected layers of a given neural network (the first approximation)}
\author{Oleg I.Berngardt}

\maketitle

\begin{abstract}%
This paper presents an algorithm for searching for the minimum number
of neurons in fully connected layers of an arbitrary network solving
given problem, which does not require multiple training of the network
with different number of neurons. The algorithm is based at training
the initial wide network using the cross-validation method over at
least two folds. Then by using truncated singular value decomposition
autoencoder inserted after the studied layer of trained network we
search the minimum number of neurons in inference only mode of the
network.

It is shown that the minimum number of neurons in a fully connected
layer could be interpreted not as network hyperparameter associated
with the other hyperparameters of the network, but as internal (latent)
property of the solution, determined by the network architecture,
the training dataset, layer position, and the quality metric used.
So the minimum number of neurons can be estimated for each hidden
fully connected layer independently. The proposed algorithm is the
first approximation for estimating the minimum number of neurons in
the layer, since, on the one hand, the algorithm does not guarantee
that a neural network with the found number of neurons can be trained
to the required quality, and on the other hand, it searches for the
minimum number of neurons in a limited class of possible solutions.

The solution was tested on several datasets in classification and
regression problems.
\end{abstract}


\section{Introduction}

In machine learning, the question often arises: how many neurons should
be in a fully connected layer of a neural network so that it can correctly
solve a given problem? For fully connected networks, there are general
theorems that allow one to answer this question: initially
\cite{Kolmogorov_1957,Arnold_1963} show that for two-layer network
it is $2N_{in}+1$ for the second layer (where $N_{in}$
is input data dimension, $N_{out}=1$ is output data dimension). 
The most recent papers improve this estimate to the following: for
an infinitely deep network in \cite{GRIPPENBERG_2003} obtained the
limit $N_{in}+N_{out}+2$; in \cite{2017arXiv171011278H} it is obtained
the limit $N_{in}+N_{out}$ for ReLU activations; in \cite{MinWidth} it is found $max(N_{in}+1,N_{out})$
limit for ReLU activations;  in most recent \cite{2022arXiv220911395C}
it is found $max(N_{in},N_{out})$ limit for any activations. 

However,
this minimum number is task-dependent and for every specific problem can be smaller than estimates above. 
So researchers are sometimes interested, for example in \cite{RUPPERT2022,Arifin_2019},
in the answer to a more specific question: how many neurons should
be in the layer of their neural network of a given architecture and
depth that solves the problem they set? In \cite{Murata} it is presented
network information criteria (NIC) to compare two trained networks
- if it is better to increase/decrease number of neurons or not. 
So one of the widely used methods to answer this question is to search
for the number of neurons as a hyperparameter found by multiple training
of the network with different number of neurons. Obviously, this approach
is resource-intensive when training complex neural networks at large
datasets.

Let's consider a method for estimating the minimum required number
of neurons in the fully connected layers of a neural network of a
given architecture solving a given problem, which do not require multiple
training the network with different number of neurons in the layers.

\section{Deterministic view}

Let us have a two-layer fully connected neural network $S$ (source)
that produces targets $T=f^{(S)}(X)$ on the dataset samples $X$:
\begin{equation}
\varphi^{\circ(1)}(W_{j,k}^{(1)}X_{i,j}+B_{k}^{(1)})=Y_{i,k}\label{eq:neuroK-1}
\end{equation}

\begin{equation}
\varphi^{\circ(2)}(W_{k,l}^{(2)}Y_{i,k}+B_{l}^{(2)})=T_{i,l}\label{eq:neuroK-2}
\end{equation}

where $X\in R^{I\times J},T\in R^{I\times L},Y\in R^{I\times K}$;
$\varphi^{\circ(i)},W_{j,k}^{(i)},B_{k}^{(i)}$ - element-wise function
of activation, weight and bias of neurons of the i-th layer, respectively;
$X_{i,j}Y_{i,k}$ - vectors of input features for the first and second
layers, respectively; $K,L$ - number of neurons in the first and
the second layer, respectively, and the number of samples $I$ in the dataset
 is greater than the number of neurons in the first layer: $I>K$.

We also have network $D$ (destination) of identical architecture,
producing results $T'=f^{(D)}(X)$ at the same dataset $X$, and differing
from network $S$ only in the number of neurons in the first layer,
as well as weights and biases $W,B$:

\begin{equation}
\varphi^{\circ(1)}(W{}_{j,m}^{('1)}X_{i,j}+B_{m}^{('1)})=Y'_{i,m}\label{eq:neuroM-1}
\end{equation}

\begin{equation}
\varphi^{\circ(2)}(W_{m,l}^{('2)}Y'_{i,m}+B_{l}^{('2)})=T'_{i,l}\label{eq:neuroM-2}
\end{equation}

where $Y'\in R^{I\times M}$, $M$ is the number of neurons in the
first layer, $T'\in R^{I\times L}$.

{\bf Definition 1\label{def:1}} (Equivalence of two networks). We
call two fully connected neural networks of the same depth and same input and output dimensions 
to be equivalent at dataset $X$,
if on each sample of the dataset $X$ they produce identical output
values $T\equiv T'$ and can differ only in the number of neurons
in each hidden layer and neuron coefficients W,B.

{\bf Lemma 1.\label{lm:1}} {\it(Equivalence of two-layer networks)  Let
the results of passing through the first fully connected layer of
two-layer fully connected networks $S$ and $D$ ((\ref{eq:neuroK-1})
and (\ref{eq:neuroM-1}) respectively) at the dataset $X$ satisfy
the linear bijection:
\begin{equation}
Y_{i,k}=Y'_{i,m}A_{m,k}\label{eq:YKtoYM}
\end{equation}
\begin{equation}
Y'_{i,m}=Y_{i,k}A_{m,k}^{(1)}\label{eq:YKtoYM-1}
\end{equation}
where $A,A^{(1)}$ are certain matrices, and $Y_{i,k}$ and $Y'_{i,m}$
are the outputs of the first layer of the networks $S$ and $D$,
respectively.
Then these networks becomes equivalent at the dataset $X$ with the
appropriate choice of coefficients of the second layer of one of the
networks.} 

\noindent
{\bf Proof.}
\label{prof:1} If the conditions (\ref{eq:YKtoYM}-\ref{eq:YKtoYM-1})
are fulfilled, the second layer of the network $D$ is:
\begin{equation}
\varphi^{\circ(2)}(W_{m,l}^{('2)}Y_{i,k}A_{m,k}^{(1)}+B_{l}^{('2)})=T'_{i,l}\label{eq:neuroM-2-1}
\end{equation}
This equation coincides with the equation (\ref{eq:neuroK-1}) when
substituting
\begin{equation}
W_{k,l}^{(2)}=W_{m,l}^{('2)}A_{m,k}^{(1)}\label{eq:W-coefs-M-1}
\end{equation}
\begin{equation}
B_{l}^{(2)}=B_{l}^{('2)}\label{eq:W-coefs-M-2}
\end{equation}
therefore, the forecasts of network $S$ on dataset $X$ with such
a choice of coefficients of the second layer repeat the forecasts
of network $D$ and $T'\equiv T$. The ability of network $D$ to
repeat the forecasts of network $S$ is proven in a similar way, therefore
$S$ and $D$ are equivalent. QED. \hfill

{\bf Corollary 1.1\label{cor:1.1}} {Under the limitations (\ref{eq:YKtoYM}-\ref{eq:YKtoYM-1})
of Lemma 1, the number of neurons in the first layer of
a fully connected two-layer neural network can be changed independently
on the number of neurons in the next layer, without changing the predicted
values. The coefficients of the neurons of the second layers can be
calculated without additional training of the new network if the bijection
transformation (\ref{eq:YKtoYM}-\ref{eq:YKtoYM-1}) matrices $A_{m,k},A_{m,k}^{(1)}$
are known.}

{\bf Lemma 2.\label{lm:2}}{\it(Minimum number of neurons in the first layer of a two-layer
equivalent network)
The minimum possible number of neurons $M$ in the first layer of
the network among all networks $D$, equivalent to network $S$ on
dataset $X$ and satisfying the limitations of Lemma 1 (\ref{eq:YKtoYM}-\ref{eq:YKtoYM-1}),
is not less than the rank of matrix $Y$ on dataset $X$.
\begin{equation}
min(M)\geq rank(Y)\label{eq:Mrank-1}
\end{equation}
} 

{\bf Proof.\label{prof:2}}
According to Lemma 1, if limitations (\ref{eq:YKtoYM}-\ref{eq:YKtoYM-1})
are satisfied, then the networks $S$ and $D$ are equivalent.
For the existence of transformations (\ref{eq:YKtoYM}-\ref{eq:YKtoYM-1}),
the ranks of the matrices $Y,Y'$ should coincide:
\begin{equation}
rank(Y)=rank(Y')\label{eq:rankYYprim1}
\end{equation}
If 
\begin{equation}
rank(Y)\neq rank(Y')\label{eq:rankYYprim2}
\end{equation}
then the networks do not satisfy the initial limitations of Lemma 1. 
The $rank(Y')$ cannot exceed the number of neurons in this layer
(the number of columns of the matrix $Y'$):
\[
M\geq rank(Y')
\]
Therefore $min(M)\geq rank(Y')$, and sequently (\ref{eq:Mrank-1}).
QED. 

{\bf Corollary 2.1 \label{cor:2.1}} {\it If among all networks satisfying
the limitations of Lemma 2 there is a network equivalent
to a given one with the number of neurons in the first layer $M=rank(Y)$,
then this is the network with the smallest possible number of neurons
in this layer among all these networks.}

{\bf Corollary 2.2 \label{cor:2.2}}
{\it
Having a trained two-layer network, under the limitations of Lemma 2 it is possible to find the lower bound of the minimum
number of neurons in the first layer of its equivalent network (\ref{eq:Mrank-1}),
and this search does not require multiple training equivalent networks
again.}

{\bf Lemma 3.\label{lm:3}}{\it (the minimum number of neurons in the hidden layers
of a multilayer fully-connected network).
Let us have a fully connected neural network with the number of hidden
layers $N$, and the outputs of n-th layer $Y^{(n)}=Y_{i,j}^{(n)}$
at the dataset $X$, $n\in[1,N]$. Then any network equivalent to
it at the dataset $X$, having $M_{n}$ neurons in n-th hidden layer
and satisfying the limitations (\ref{eq:YKtoYM}-\ref{eq:YKtoYM-1})
of Lemma 1 in each hidden layer also satisfies the condition:
\begin{equation}
M_{n}\geq rank(Y^{(n)})\label{eq:Mrank}
\end{equation}
} 

{\bf Proof. \label{prof:3}}
Let's consider layers $n$ and $n+1$ as a separate fully connected
neural network with inputs $Y^{(n-1)}$ and outputs $Y^{(n+1)}$.
A neural network equivalent to it satisfying the limitations of Lemma 1
 and having the minimum possible number of neurons in this
layer has at least $rank(Y^{(n)})$ neurons in the first layer (corresponding
to layer n of the original network) in accordance with Lemma 2.
This is true for each layer except the last layer of the original
network, i.e. for all hidden layers of the original network ($n\leq N$).
QED.

{\bf Corollary 3.1 \label{cor:3.1}}
{\it If a network $D$ with $rank(Y^{(n)})$ neurons in layer $n$ is equivalent
to some network $S$ with $Y^{(n)}$ outputs at dataset $X$, then it has the minimum possible
number of neurons in this layer among all the equivalent to $S$ networks
at this dataset that satisfy the limitations (\ref{eq:YKtoYM}-\ref{eq:YKtoYM-1})
of Lemma 1.}

\section{Stochastic view}

Based on Lemma 3, the problem of finding the minimum number
of neurons in each hidden layer ($n$) of a fully connected neural
network can be reduced to search for the $rank(Y^{(n)})$ : rank of
layer outputs at this dataset with following single retraining of
the resulting network with found minimal number of neurons to finally
check its equivalence to the original network.

Let's assume that we already have trained fully connected neural network
$S$ of sufficiently large width that solves some problem $X\rightarrow T$,
$X\in R^{I\times J};T\in R^{I\times L}$. Then we can find the minimum
number of neurons in the equivalent network $D$, for each hidden
fully-connected layer separately.

When searching, one should to take into account the following facts:
\begin{itemize}
\item Training a neural network is a stochastic process, and the results
and coefficients of neurons are determined by a combination of random
factors: initial conditions and the learning process. Therefore, the
matrices $Y^{(n)}$ obtained during each training of the same network
are different, and may have different ranks: for $C$ independent
trainings of the same $S$ network, instead of the matrix $Y^{(n)}$
we have a set of $C$ matrices $\{Y^{(n)}\}_{C}$, with ranks $\{M^{(n)}\}_{C}$
accordingly;
\item Taking into account the limited accuracy of numerical algorithms and
source data, finding $rank(Y^{(n)})$ (and the matrix $Y^{(n)}$ turns
out to be very large) is associated with calculating its truncated
SVD decomposition and choosing the optimal level $M$ of this truncat.
For large matrices, solving this problem is usually difficult, and
there are many ways to choose the truncat level $M$ for truncated
SVD decomposition (corresponding to the rank of the matrix). But these
ways produce different values, so the choice of $M$ is subjective
problem, as shown by \cite{PERESNETO2005,Bai2002,FALINI2022} and therefore depends
on the actual problem we solve by the network.
\end{itemize}
Thus, the task of constructing a minimal network $D$ equivalent to
a network $S$ with a given architecture should be solved from a statistical
point of view and could be subjective. A similar remark is true about
minimum number of neurons and networks equivalence.

{\bf Definition 2 \label{def:2}}(statistical equivalence of networks). 
Let's call two networks $S$ and $D$ statistically equivalent at
a dataset $X$ with a quality metric $Q$ and its threshold level
$Q_{0}$ ($Q_{0}$ worse than absolutely the best value of $Q$),
if the networks $S$ and $D$ at the dataset $X$ produce almost identical
forecasts - with the quality metric $Q$ not worse than $Q_{0}$.

{\bf Corollary 4.1 \label{col:d2.1}} Network $S$ is statistically
equivalent to itself with any pre-specified threshold level $Q_{0}$
satisfying the limitations of the Definition 2.

{\bf Corollary 4.2\label{col:d2.2}}  All fully overtrained (completely
repeating the targets/labels $T$ of training dataset $X$) networks
on a given dataset $X$ are equivalent and statistically equivalent
at this dataset with any pre-specified threshold level $Q_{0}$ satisfying
the limitations of the Definition 2.

Let us formulate an algorithm for finding minimum number of neurons
in fully connected layer by checking the statistical equivalence of
two networks - wide one $S$ and probe one $D$.

To obtain a set of matrices $\{Y\}_{C}$, we use cross-validation
- dividing the original dataset $X$ into $C$ parts. We use randomly
selected $C-1$ parts of the dataset for training, and the rest of
the dataset to validate and check the stopping conditions of training.
Thus, we get $C$ variants of $S$ network, and $C$ variants of matrix
$Y$ (set of matrices $\{Y\}_{C}$).

From Corollary 4.2 it follows that checking the statistical
equivalence of two networks should be carried out on a dataset, part
of which none of the networks were trained on. At the part of the
dataset on which they were both trained, they can be equivalent with
any predefined quality $Q_{0}$ (satisfying the limitations of the
Definition 2 ) simply due to their overtraining.

Therefore as a dataset $X'$ for checking statistical equivalence,
we choose a dataset, half of the elements of which were used to train
network $S$, and another half - to validate network $S$. 

Taking into account the property of most quality metrics at non-intersecting
datasets $X_{1},X_{2}$:

\begin{equation}
Q(X_{1}\vee X_{2})=\frac{Dim(X_{1})Q(X_{1})+Dim(X_{2})Q(X_{2})}{Dim(X_{1})+Dim(X_{2})};X_{1}\wedge X_{2}=\emptyset\label{eq:Qprops}
\end{equation}

, equality of fold sizes ($Dim(X_{1})\approx Dim(X_{2})$), and Corollary~4.2, 
we choose the threshold $Q_{0}$ equal to the average
value between the quality metric $Q$ of network $S$ at the validation
fold $X_{1}$ ($\overline{Q_{@val}}(S)$) and the best possible quality
metric $Best(Q)$ - at training fold $X_{2}$.

\begin{equation}
Q_{0}=\frac{\overline{Q_{@val}}(S)+Best(Q)}{2}\label{eq:Q0}
\end{equation}

From each network $S_{C_{i}}$ we produce a network $D_{C_{i}}$,
and check these two networks for statistical equivalence with threshold
level (\ref{eq:Q0}). Since each network $S,D$ produce different
results depending on the training dataset and on the dataset on which
the comparison is made, it is possible to obtain $C(C-1)$ estimates
of the network quality metric $Q_{ij}=Q(S_{C_{i}}(X'_{ij}),D_{C_{i}}(X'_{ij}));i\neq j$
where checking dataset $X'_{ij}$ is: 
\begin{equation}
X'_{ij}=(X\setminus X^{(C_{i})})\lor(X\setminus X^{(C_{j})})\label{eq:XcrossVal}
\end{equation}

where $X^{(C_{i})},X^{(C_{j})}$ - training datasets used for i-th
and j-th cross-validation.

Quality is a stochastic quantity, so we need to know the worst bound
of the metric $Q_{ij}$ (with some significance level $\alpha$).
If this boundary is not worse than $Q_{0}$, then the networks $S,D$
considered to be statistically equivalent with threshold level $Q_{0}$. 

Unfortunately, the number of folds $C$ is usually small and traditionally
does not exceed 10, so it is difficult to reach high enough level
of statistical significance using this data only. We use traditional
way to increase the size of the ensemble - bootstrap technique, presented in \cite{Bootstrap}. 

The resulting approximate method for assessing statistical equivalence
is presented in Algorithms \ref{alg:1}-\ref{alg:2}. It estimates
the average $Q$ (in algorithm $\overline{Q_{@val}}$) based on the
minimum $Q$ obtained during bootstrap estimate of $Q$ for each of
the $X_{ij}$ datasets. The quality estimate is shown in Algorithm
\ref{alg:1}. 

\begin{algorithm}
\caption{Estimating the quality metric between model predictions}
\label{alg:1} \begin{algorithmic}

\Procedure {WorstQ} {$X,X_{C},S_{C},D_{C},Q,i,j$}

\hspace*{\algorithmicindent} \textbf{Input}: data set X

\hspace*{\algorithmicindent} \textbf{Input}: C training folds $X^{(C_{i})}$

\hspace*{\algorithmicindent} \textbf{Input}: C trained networks $S_{C_{i}},D_{C_{i}}$
at folds $X^{(C_{i})}$

\hspace*{\algorithmicindent} \textbf{Input}: metric Q

\hspace*{\algorithmicindent} \textbf{Input}: cross-validation folds
numbers - i,j

\State Calculate network outputs: 

\State $T\gets S_{C_{i}}(X\setminus(X^{(C_{i})}\wedge X^{(C_{j})}))$ 

\State $T'\gets D_{C_{j}}(X\setminus(X^{(C_{i})}\wedge X^{(C_{j})}))$ 

\State $N\gets10000$ 

\State $Qset\gets\emptyset$ 

\For {$iteration\in[1..N]$} 

\State $t,t'\gets BootstrapedPairs(T,T')$ 

\State $Qset\gets Qset\vee Q(t,t')$ 

\EndFor 

\State \Return $Worst(Qset)$ 

\EndProcedure 

\end{algorithmic} 
\end{algorithm}

Comments to Algorithm \ref{alg:1}: 
\begin{itemize}
\item Function $Q(T,T')$ returns the value of the $Q$ metric between the
outputs of networks $S,D$ based on pairs of the network output forecasts
$(T_{i},T'_{i})$ ;
\item The BootstrappedPairs function returns pairs $(T_{i},T'_{i})$ randomly
selected from samples $T,T'$, the number of pairs returned is equal
to the number of samples in $T$ (and $T'$) ;
\item The $Worst(Qset)$ function determines the worst metric value from
a set $Qset$. For example, for the case $Q=Accuracy$ this is the
minimum value over the set, for $Q=MSE$ it is the maximum value over
the set;
\item Number of folds $C\ge2$ to ensure that the networks was not trained
at the entire dataset used for statistical equivalence check (see
Corollary~4.2);
\item Number of bootstrap iterations $N$ can be used to estimate the minimal
significance level of the algorithm as $\alpha\sim1/N$.
\end{itemize}
Knowing how to check the statistical equivalence of two networks,
and having a set of networks $S$ trained by cross-validation using
$C$ folds, we can formulate an algorithm for finding the minimum
number of neurons in a layer.

Since the choice of SVD truncating level $M$ is subjective, as shown by \cite{PERESNETO2005,Bai2002,FALINI2022},
the problem of finding the minimum number of neurons can be formulated
as follows: find the minimum number of neurons $M$ in network $D$
such that networks $S$ and $D$ remained statistically equivalent
with given threshold level $Q_{0}$ and quality metric $Q$.

To find the ranks $\{M\}_{C}$ of the set of matrices $\{Y\}_{C}$,
we create $D'$ network by putting a linear transformation between
the first and second fully connected layers:

\begin{equation}
Y'=YA^{(M)}A^{(M)+}\label{eq:linDprim}
\end{equation}
 where $A^{(M)},A^{(M)+}$ are the matrices of the direct and pseudo-inverse
truncated SVD transformation, both truncated at level $M$:

\begin{equation}
\varphi^{\circ(1)}(W_{j,k}^{(1)}X_{i,j}+B_{k}^{(1)})=Y_{i,k}\label{eq:neuroK-1-1}
\end{equation}

\begin{equation}
Y'_{i,r}=A_{s,r}^{(M)+}A_{k,s}^{(M)}Y_{i,k}\label{eq:neuroK-1-1-1}
\end{equation}

\begin{equation}
\varphi^{\circ(2)}(W_{r,l}^{(2)}Y'_{i,r}+B_{l}^{(2)})=T_{i,l}\label{eq:neuroK-2-1}
\end{equation}

Here $W_{j,k}^{(1)},B_{k}^{(1)},W_{r,l}^{(2)},B_{l}^{(2)}$ - weights
and biases of the trained network $S$.

For $M\geq rank(Y)$, by SVD definition, $Y'\equiv Y,T'\equiv T$
and the networks $S,D'$ are equivalent and statistically equivalent. 
Therefore,
the task of finding the rank of the matrix $Y\in\{Y\}_{C}$ is reduced
to finding the minimum $M$ for which the networks $S,D'$ are statistically
equivalent in terms of the metric $Q$ with good enough threshold
level $Q_{0}$ (Algorithm \ref{alg:1}).

\begin{figure}
\includegraphics[scale=0.6]{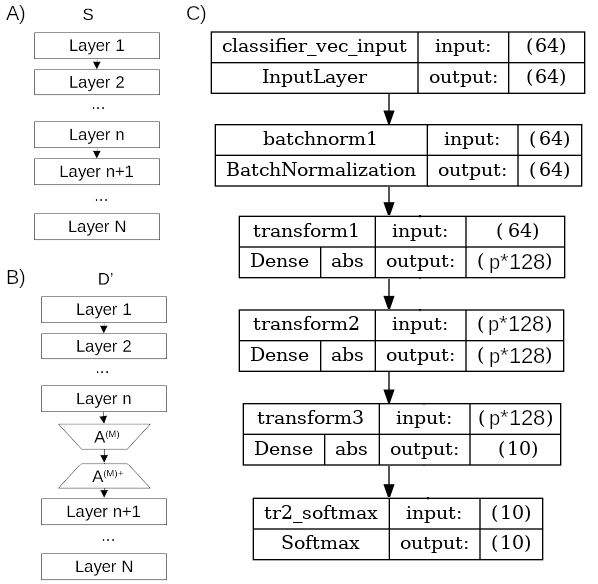}
\caption{Architecture for searching for the minimum number of neurons in a layer
$n$. A) - architecture of network $S$; B) - architecture of network
$D'$; C) - architecture for study MNIST problem, p is a network width
multiplier.}
\label{fig:SVDauto}
\end{figure}

The transformation $A^{(M)+}A^{(M)}Y$ is essentially an autoencoder-like architecture, 
first described in \cite{Autoencoders}, but
is based on the truncated SVD transformation. It does not distort the
input data when the size of its latent representation $M$ is not
less than the rank of the matrix $Y$. SVD Autoencoders has been already
used for optimizing neural networks, for example in \cite{xue2013,Xue2014},
but not in the task of finding minimum number of neurons in fully
connected layer without multiple training. The architectures of the
original network $S$ (\ref{eq:neuroK-1}-\ref{eq:neuroK-2}) and
network $D'$ (\ref{eq:neuroK-1-1}-\ref{eq:neuroK-2-1}) are shown
in Fig.\ref{fig:SVDauto}A-B.

Thus, the search for the minimum number of neurons, assuming that
the limitations of Lemma 1 are met, comes down to calculating
the quality metrics over an ensemble of already trained networks.
We put the truncated SVD autoencoder after the studied layer and study
the metrics of the forecast quality of these networks over the studied
dataset at various levels of its truncat, as shown in Algorithm \ref{alg:2}. 

\begin{algorithm}
\caption{Finding the minimum number of neurons in a layer}
\label{alg:2}

\begin{algorithmic} 

\Procedure {FindNumberOfNeurons}{$S,L,X,C,Q$}

\hspace*{\algorithmicindent} \textbf{Input}: network $S$

\hspace*{\algorithmicindent} \textbf{Input}: layer number $L$

\hspace*{\algorithmicindent} \textbf{Input}: data set $X$

\hspace*{\algorithmicindent} \textbf{Input}: folds number $C$

\hspace*{\algorithmicindent} \textbf{Input}: metric $Q$

\State Make $C$ folds $X^{(C_{i})}$ from $X$ of nearly equal size:

\State $(X\setminus X^{(C_{i})})\wedge(X\setminus X^{(C_{j\neq i})})=\emptyset,\vee_{i\in[1..C]}(X\setminus X^{(C_{i})})=X$

\For {$i\in[1..C]$} 

\State Train $S_{C_{i}}$ networks: 

\State $S_{C_{i}}\gets$ $S$ trained at $X^{(C_{i})}$, validated
at $X\setminus X^{(C_{i})}$ 

\State$Q_{i}\gets Q(S_{C_{i}})$ at $X\setminus X^{(C_{i})}$

\State Calculate layer $L$ outputs: 

\State $Y_{C_{i}}$: $Y_{C_{i}}\gets S_{C_{i}}(X,L)$ 

\State Calculate SVD decompositions of layer outputs: 

\State $U_{i}\Sigma_{i}V_{i}^{T}\gets Y_{C_{i}}$ 

\EndFor

\State $\overline{Q_{@val}}\gets Mean(\{Q_{i}\})$

\State $Q_{search}\gets\left(BestQ+\overline{Q_{@val}}\right)/2$

\State $M_{found}\gets\emptyset$

\For {$i\in[1..C]$} 

\For {$j\in[1..C],j\neq i$} 

\State $M_{min}\gets1$ 
\State $M_{max}\gets Dim(Y_{C})$

\For {$iteration\in[0,log_{2}(Dim(Y_{C})+1)]$}

\State $M\gets(M_{max}+M_{min})/2$

\For {$k\in[1..C]$} 

\State $D'_{C_{k}}\gets S_{C_{k}}$ 

\State $AutoEnc_{k}\gets makeTruncuttedSVDautencoder(\Sigma_{k},V_{k}^{T};M)$ 

\State $D'_{C_{k}}\gets InsertAfterNetworkLayerL(AutoEnc_{k},D'_{C_{k}},L)$ 

\EndFor

\State $Q'\gets WorstQ(X,X_{C},S_{C},D'_{C},Q,i,j)$

\If {$Q'\,worser\,than\,Q_{search}$}

\State $M_{min}\gets M$ 

\Else 

\State $M_{max}\gets M$ 

\EndIf

\If {$M_{max}-M_{min}<=1$} 

\State $M_{found}\gets M_{found}\vee M_{max}$ 
\State {\bf break for} 

\EndIf

\EndFor

\EndFor

\EndFor

\State \Return $Mean(M_{found})$ 

\EndProcedure

\end{algorithmic} 
\end{algorithm}

Comments to the Algorithm \ref{alg:2}.
\begin{itemize}
\item BestQ - the best possible value of the $Q$ metric: for example, for
accuracy in a classification problem it is 1 (100\%), and for MSE
or MAE in a regression problem it is 0.
\item The search for the truncat level $M$ for the SVD autoencoder is carried
out by searching for the point $Q(M)=Q_{0}$ by bisection method (which
is accurate under the assumption that the dependence $Q(M)$ is smooth
and monotonic, which is not always the case).
\item Due to Lemma 3 the algorithm \ref{alg:2} can be independently
repeated for any hidden fully connected layer of network $S$, splitting
$X$ into cross-validation folds and training the variants of networks
$S_{C_{i}}$ only once, which further reduces the time and required
resources.
\end{itemize}
In accordance with the Corollary~3.1 of Lemma 3,
after determining the minimum number of neurons by Algorithm \ref{alg:2},
it is necessary to retrain the network of architecture $S$ with the
found minimum number of neurons in layers to make sure that it is
statistically equivalent to the original networks $S_{C_{i}}$ with
threshold level $Q_{0}=Q_{@val}$ at test dataset.

\section{Experiments}

To test the performance and stability of the algorithm, the MNIST
dataset, presented in \cite{Lenet5} and the classification task were chosen. Datasets
of training single-color 8x8 images were studied. The dataset has
10 classes. We used the MNIST dataset variant from the Tensorflow
library (28x28 images, 60000 training, 10000 test images), the images
were reduced to 8x8 size to speed up calculations (MaxPooling + Padding),
and flattened (reshaped) into 64-dimension vector.

For experiments, a neural network with three layers was used - two
hidden layers of equal (variable) width, an output layer of 10 neurons,
and an output Softmax activation. The input has a batch normalization
layer. Activation function of all the layers - Abs. 

The network architecture is shown in Fig.\ref{fig:SVDauto}C. The
network has formula: Softmax, FCx(10)(Abs), FCx(p{*}128)(Abs), FCx(p{*}128)(Abs),
BN where $p$ is some layer width multiplier, and BN - Batch Normalization
layer, FC - Fully connected layer. The SVD autoencoder was implemented
on TensorFlow through two layers of untrained embeddings, the coefficients
of which were set from the truncated SVD decomposition matrices (direct
and inverse). For training we used ADAM optimizer with constant learning
rate $10^{-3}$, early stopping with patience 3 and monitoring of
loss function at validation. Loss function is cross-entropy, quality
metric $Q$ is Accuracy. 

During the experiments, the following criteria for the stability and
performance of the algorithm were numerically checked:
\begin{enumerate}
\item Stability of the result to changes in the number of neurons in the
original layer (variants with different $p$ in Fig.\ref{fig:SVDauto}C);
\item Stability of the result to the choice of a combination of folds for
comparison: pairs $i,j$ on cross-validation;
\item Stability of the result to changes in the number of folds (variants
for different $C\in[2..7]$);
\item Non-decreasing forecast quality (forecast quality of found network
is not worse than of original network);
\item Equivalence of the original network and new one at the test dataset.
\end{enumerate}
The results of estimating the optimal number of neurons for each width
(by varying elastics coefficient $p$ in network $S$, Fig.\ref{fig:SVDauto}C)
and for each layer are marked in Fig.\ref{fig:SVDauto}C as transform1
and transform2. It can be seen from the figure that the method reaches
a constant value when the ratio between the number of neurons in layer
$S$ and the predicted minimum value in layer $D'$ is greater than
approximately 3. As one can see the minimum width of the first layer
is about 40 neurons, and the minimum width of second layer is about 10 neurons.

\begin{figure}
\includegraphics[scale=0.5]{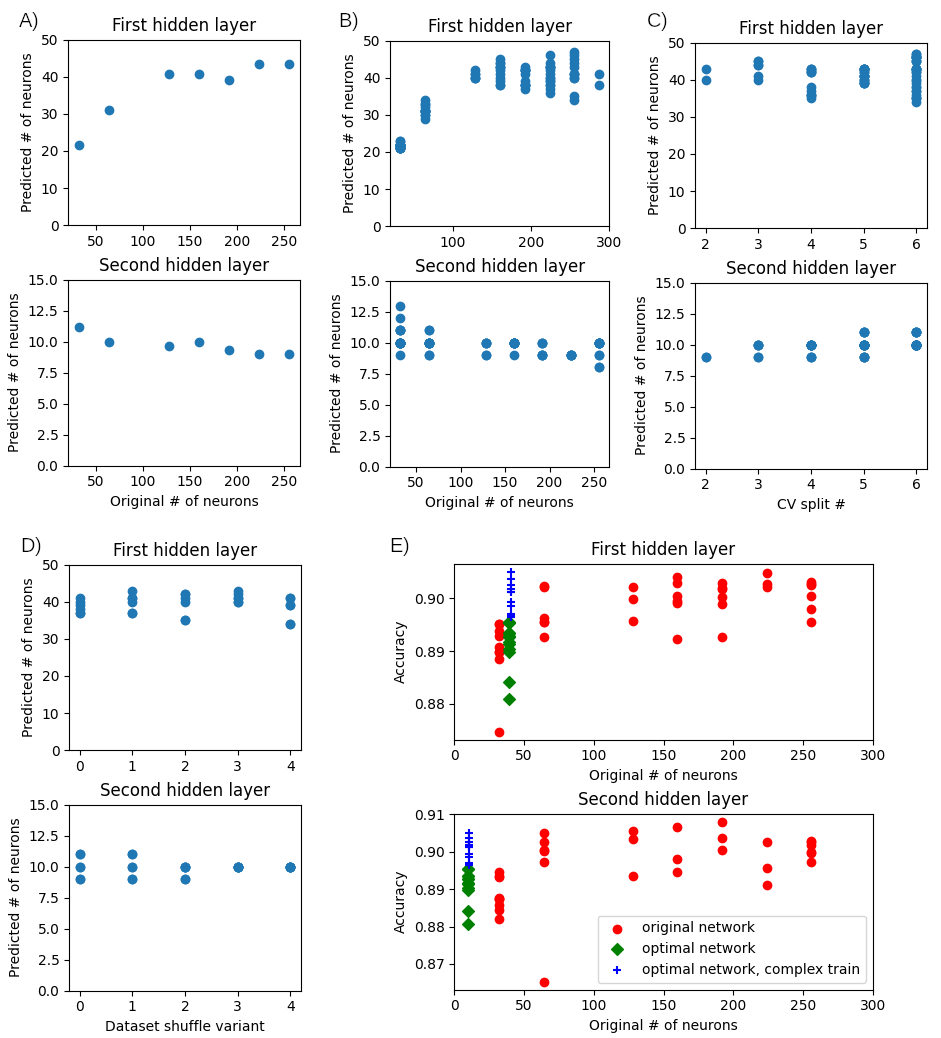}
\caption{Performance and stability of the algorithm. A) predicted minimum number
of neurons for different number of neurons in the $S$ original layer
(mean over $C$ folds combinations); B) predicted minimum number of
neurons for different number of neurons in the $S$ original layer
(over $C$ folds combinations); C) predicted minimum number of neurons
for different number of $C$ folds (for the number of neurons in original
layer is 128); D) predicted minimum number of neurons for different
dataset separation and order variants (for the number of neurons in
original layer is 128); E) accuracy at test dataset for original network
$S$ (red circles) and found equivalent network $D$ with minimal
number of neurons in the layers (40 and 10 for the first and second
layer correspondingly): green diamonds - for simple training with
constant learning rate $10^{-3}$and early stopping with patience
3; blue crosses - for decreasing learning rate $10^{-3}-10^{-6}$
and early stopping with patience 10}
\label{fig:perform}
\end{figure}

Fig.\ref{fig:perform}B shows the results of testing robustness to
the choice of fold pair for comparison. In Algorithm \ref{alg:2}
these values are stored in the array $M_{found}$ and are averaged
to obtain the result for a given network width. The Fig.\ref{fig:perform}B
shows the data before averaging over pairs of folds. The figure shows
that the spread of values across fold pairs is relatively low and
on average does not exceed 10-20\%, which, when averaging over pairs
of folds, makes the standard deviation several times less.

Fig.\ref{fig:perform}C shows the dependence of the predicted number
of neurons as a function of the number of folds $C$ during cross-validation
(before averaging over the pairs of folds). It can be seen from the
figure that the dependence on the number of folds $C$ is weak and
therefore even cross-validation over 2 or 3 folds can be used to solve
the problem.

Fig.\ref{fig:perform}D shows the results of testing the stability
of the algorithm to the cross-validation partitioning variants, the
width of the original network is 128 neurons. Previously, when studying
the algorithm, we set the samples order in the training dataset in
a fixed way; the figure shows how the results depends on the mixing
the order, and shows the distribution of the results for each pair
of folds. The figure shows that the results are stable with an accuracy
of about 10\%. Thus, the result depends little on the order of elements
in the dataset and seperation to the cross-validation folds.

After this, the neural network was trained 10 times with the minimum
number of neurons found (40 neurons in the first layer - transform1,
10 in the second layer - transform2), the obtained accuracy at the
test dataset are shown in Fig.\ref{fig:perform}E. The figure shows
that the achieved accuracies (shown in Fig.\ref{fig:perform}E with
diamonds) with standard training are slightly worse (the intervals
intersect, but the averages are separated) from the accuracies achieved
by the original networks. However, with more complex training (decreasing
learning rate from $10^{-3}$ to $10^{-6}$ with step 0.1, increasing
patience of early stopping from 3 to 10), the achieved accuracies
becomes nearly the same with original network (shown in Fig.\ref{fig:perform}E
with crosses).

For the final check of the equivalence of the found network to the
original ones, quality metrics were calculated between the prediction
results of three original networks S with a width of 128 neurons of
each hidden layer and three found networks D with a width of 40 and
10 neurons of the corresponding hidden layers. Three variant of networks
in each case were obtained by cross-validation. The resulting 9 quality
metric values (accuracy) are in the range 0.919..0.931. The mutual
quality of the networks at the test dataset is not lower than the
quality of the original network at the test dataset, which suggests
that found network (D) with a minimum number of neurons is equivalent
to the original one (S).

\section{Discussion}

The presented results show that the proposed algorithm is stable enough
on average: on average, the estimated minimum layer width weakly depends
on the initial number of neurons in the original layer when the neural
network is wider than 3 minimum widths. The minimum number of neurons
can indeed be found for each layer independently, which suggests that
the number of neurons is not a hyperparameter, the combination of
which determines the accuracy of the solution, but an internal property
determined for each layer separately. This suggests that on average
we actually find some latent dimension of the layer, which is the
basic internal property of the solution, and this does not depend
on how wide the original layer width is. Therefore, to find the internal
(latent) dimension, it is enough for us to train a fairly wide network
with the number of neurons exceeding the expected minimum number of
neurons by at least 3 times; with a smaller width of the original
network, the resulting forecast could be inaccurate.

We also tested the algorithm on other problems - both classification
and regression: fullsize 28x28 MNIST, FashionMNIST dataset, presented in \cite{xiao2017}
(classification task), as well as California housing, shown in \cite{CalHousing}
and Wine Quality demonstrated in \cite{WineQ} datasets (regression task). In the first
case, a network with two hidden layers was used, in other cases -
with one hidden layer. In all the cases, the initial width of the
hidden layers was chosen to be 200 neurons (and 300 for MNIST task),
and cross-validation was carried out using 3 folds. The result are
shown in Table \ref{tab:others}. The table shows that the solutions
found are equivalent to original ones with $Q_{0}$ not worse than
$Q_{@val}$, so the algorithm looks useful. As one can also see, the
found minimum number of neurons is much smaller than expected both
from \cite{GRIPPENBERG_2003} ($N_{in}+N_{out}+2$) universal formula
predictions (where $N_{in},N_{out}$ are input and output dimensions
respectively), and from \cite{2022arXiv220911395C} universal formula
predictions ($max(N_{in},N_{out})$). 

It is important to note that discussed approach can be useful not
only for fully connected networks, but for any network architecture 
to estimate the minimal size of any fully connected layer followed 
by another fully connected one. 

\begin{table}
\caption{Algorithm results for other tasks, C=3. Designations: BN - Batch Normalization
layer, FC - Fully connected layer, FL - Flatten layer. }

\begin{tabular}{|p{3.5cm}|p{2.5cm}|p{2cm}|p{2cm}|p{2cm}|}
\hline 
\textbf{Data set} & MNIST 28x28 & Fashion MNIST & Calfornia housing & Wine Quality\tabularnewline
\hline 
\hline 
\textbf{Task} & \multicolumn{2}{c|}{Classification} & \multicolumn{2}{c|}{Regression}\tabularnewline
\hline 
\textbf{Metric }$Q$ & \multicolumn{2}{c|}{Accuracy} & \multicolumn{2}{c|}{MSE}\tabularnewline
\hline 
\textbf{Source network }$S$\textbf{ formula} & FCx10 (Softmax),

FCx300 (ReLU), FCx300 (ReLU), BN, FL & FCx10 (Softmax),

FCx200 (ReLU), FCx200 (ReLU), BN, FL & \multicolumn{2}{c|}{FCx1(linear), FCx200(ReLU), BN}\tabularnewline
\hline 
\textbf{$S_{C_{i}}$ metric value at validation dataset} & 0.926..0.950 & 0.883..0.887 & 0.442..0.490 & 0.476..0.559\tabularnewline
\hline 
\textbf{Resulting network}

$D$\textbf{ formula} & FCx10 (Softmax),

FCx19 (ReLU), FCx68 (ReLU), BN, FL & FCx10 (Softmax),

FCx25 (ReLU), FCx36 (ReLU), BN, FL & FCx1(linear), FCx5(ReLU), BN & FCx1(linear), FCx6(ReLU), BN\tabularnewline
\hline 
\textbf{$D_{C_{i}}$ metric value at validation dataset} & 0.962..0.964 & 0.886..0.888 & 0.516..0.521 & 0.500..0.621\tabularnewline
\hline 
\textbf{Metric value between $S_{C_{i}}$ and $D_{C_{i}}$ at test
dataset} & 0.922..0.941 & 0.869..0.894 & 0.049..0.084 & 0.060..0.088\tabularnewline
\hline 
\textbf{Found minimum number of neurons in layers by our algorithm} & 68, 19 & 36, 25 & 5 & 6\tabularnewline
\hline 
\textbf{Expected minimum number of neurons per layer from \cite{GRIPPENBERG_2003}} & 796, 796 & 796, 796 & 11 & 14\tabularnewline
\hline 
\textbf{Expected minimum number of neurons per layer from \cite{2022arXiv220911395C}} & 784, 784 & 784, 784 & 8 & 11\tabularnewline
\hline 
\end{tabular}\label{tab:others}
\end{table}

The main problems of this algorithm are:
\begin{enumerate}
\item The algorithm is stochastic, so the minimal number of neurons it predicts
is also stochastic, so one cannot guarantee that the minimum number
of neurons cannot be greater or smaller than the number found by this
algorithm.
\item The algorithm is computationally expensive at the initial stage -
it requires $C$ trained versions of the original network by cross-validation
using $C$ folds, but this is often a standard procedure for estimating
the accuracy of the network. It was shown that 2-3 folds looks enough
for calculations.
\item The difficulty of creating the initial complete SVD decomposition
of the outputs of the neural layer. To speed things up, one can try
to get it from random $N>J$ samples of the dataset (where $J$ is
the number of output neurons of the studied layer of network $S$)
but with accuracy loss;
\item Cumbersome calculations by truncated SVD autoencoder - it uses matrix
multiplications; in addition, it is necessary to calculate quality
metrics between variants of $S$,$D'$ at different fold combinations,
which also slows down the algorithm. 
\item The algorithm operates under the assumption that the limitations of
Lemma 1 are met. Due to the nonlinearity of the activation
functions, it is not obvious that an equivalent network with a minimum
number of layer's neurons satisfies the limitations of Lemma 1,
so one cannot guarantee that the minimum number of neurons cannot
be less than the number found by this algorithm. 
\item It is not obvious how (and whether it is possible) to formalize the
calculation of the minimum number of neurons for the case of non-elementwise
activation functions, for example Softmax or Maxout: in this case,
activation functions can create additional relationships between the
columns of the matrix $Y$ and change the rank of the output matrix
compared to the rank of the argument matrix. In the case of Softmax,
for example, the rank of the output matrix is 1 less than the number
of neurons due to the normalization process.
\item It is not obvious how (or if) the algorithm will work in the case
of training with random data augmentation, widely used now when training
networks.
\end{enumerate}

\section{Conclusion}

The paper presents an algorithm for searching of the minimum number
of neurons in a fully connected layers of a network. The basis of
the algorithm is training the initial wide network using the cross-validation
method using at least two folds. After training, for analyzed fully
connected layer, truncated SVD autoencoders are built for each trained
network. Each SVD autoencoder is inserted inside the corresponding
network after the studied layer. The quality of the resulting ensemble
of networks is determined by comparing the forecasts of the original
network and the network with a truncated SVD autoencoder, on a dataset
composed of a pair of folds of the original dataset. The statistical
equivalence of new and original networks is determined by the quality
metric reaching a threshold value - the average between the metric
of the original network at the validation dataset and the best possible
metric value. The algorithm constructed in this way searches for the
minimum dimension of the truncated SVD autoencoder, which has the
meaning of the rank of the matrix of output values of the layer of
the original network and, within the framework of the described approach,
corresponds to the minimum number of neurons in the fully connected
layer.

The minimum number of neurons in a layer determined by this algorithm
does not require multiple training of the network for different values
of the number of neurons in the layer. Therefore minimum number of
neurons is not a hyperparameter related with other hyperparameters
of the network, but an internal property of the solution, and can
be calculated independently for each layer, and depends on given architecture
of the network, layer position, quality metric, and training dataset. 
The proposed algorithm determines
the first approximation for estimating the minimum number of neurons,
since on the one hand it does not guarantee that a neural network
with the found number of neurons can be trained to the required quality
(which will ultimately lead to the increase of the found minimum number
of neurons), and on the other hand, it searches for the minimum number
of neurons in a limited class of solutions that satisfies the limitations
of Lemma 1 (which ultimately lead to the decrease of the
found minimum number of neurons).

An experimental study of the algorithm and properties of the solution
was carried out on the smallsize (8x8) MNIST dataset, and demonstrated
the effectiveness of the proposed solution in this problem. The described
algorithm was also tested on several well-known classification and
regression problems, including full 28x28 MNIST images.

Sample code is avaliable at https://github.com/berng/FCLayerMinimumNeuronsFinder .

\section*{Acknowledgments}
The work has been done under financial support of the Russian Science
Foundation (grant No.24-22-00436).

\vskip 0.2in
\bibliographystyle{unsrt}
\bibliography{references}

\end{document}